\def\year{2016}\relax
\documentclass[letterpaper]{article}
\usepackage{aaai16}
\nocopyright{}
\usepackage{times}
\usepackage{helvet}
\usepackage{courier}

\usepackage{algorithm}
\usepackage{algorithmic}
\usepackage{amsmath}
\usepackage{amssymb}
\usepackage{amsthm}
\usepackage{graphicx}
\usepackage{appendix}

\newcommand{\Algorithm}{Adaptive $\lambda$ Least-Squares Temporal Difference learning}
\newcommand{\Alg}{ALLSTD}

\newtheorem{define}{Definition}
\newtheorem{theorem}{Theorem}

\begin{document}
\title{Adaptive $\lambda$ Least-Squares Temporal Difference Learning}
%%%%%%%%%%%%%%%%%%%%%%%%%%
\author{Timothy A.~Mann \and Hugo Penedones \and Todd Hester\\
Google DeepMind\\
London, United Kingdom\\
\{kingtim, hugopen, toddhester\}@google.com\\
\And
Shie Mannor\\
Electrical Engineering\\
The Technion\\
Haifa, Israel\\
shie@ee.technion.ac.il
}
\maketitle
\begin{abstract}
  Temporal Difference learning or TD($\lambda$) is a fundamental algorithm in the field of reinforcement learning. However, setting TD's $\lambda$ parameter, which controls the timescale of TD updates, is generally left up to the practitioner. We formalize the $\lambda$ selection problem as a bias-variance trade-off where the solution is the value of $\lambda$ that leads to the smallest Mean Squared Value Error (MSVE). To solve this trade-off we suggest applying Leave-One-Trajectory-Out Cross-Validation (LOTO-CV) to search the space of $\lambda$ values. Unfortunately, this approach is too computationally expensive for most practical applications. For Least Squares TD (LSTD) we show that LOTO-CV can be implemented efficiently to automatically tune $\lambda$ and apply function optimization methods to efficiently search the space of $\lambda$ values. The resulting algorithm, \Alg is parameter free and our experiments demonstrate that \Alg\ is significantly computationally faster than the na\"{i}ve LOTO-CV implementation while achieving similar performance.
\end{abstract}

\noindent The problem of policy evaluation is important in industrial applications where accurately measuring the performance of an existing production system can lead to large gains (e.g., recommender systems \cite{Shani2011}). Temporal Difference learning or TD($\lambda$) is a fundamental policy evaluation algorithm derived in the context of Reinforcement Learning (RL). Variants of TD are used in SARSA \cite{Sutton1998}, LSPI \cite{Lagoudakis2003}, DQN \cite{Mnih2015}, and many other popular RL algorithms.

The TD($\lambda$) algorithm estimates the value function for a policy and is parameterized by $\lambda \in [0, 1]$, which averages estimates of the value function over future timesteps.
% However, the weight assigned to later timesteps decreases as $\lambda \rightarrow 0$ and increases as $\lambda \rightarrow 1$. Thus, $\lambda$ controls the timescale of TD. Setting $\lambda$ close to 0 may introduce bias, but generally results in small variance updates. On the other hand, setting $\lambda$ close to 1 minimizes bias but tends to result in high variance updates. Thus,
 The $\lambda$ induces a bias-variance trade-off. Even though tuning $\lambda$ can have significant impact on performance, previous work has generally left the problem of tuning $\lambda$ up to the practitioner (with the notable exception of \cite{White2016}). In this paper, we consider the problem of automatically tuning $\lambda$ in a data-driven way.

\noindent {\bf Defining the Problem:} The first step is defining what we mean by the ``best'' choice for $\lambda$. We take the $\lambda$ value that minimizes the MSVE as the solution to the bias-variance trade-off.

 \noindent {\bf Proposed Solution:} An intuitive approach is to estimate MSE for a finite set $\Lambda \subset [0, 1]$ and chose the $\lambda \in \Lambda$ that minimizes an estimate of MSE.
 \noindent {\bf Score Values in $\Lambda$:} We could estimate the MSE with the loss on the training set, but the scores can be misleading due to overfitting. An alternative approach would be to estimate the MSE for each $\lambda \in \Lambda$ via Cross Validation (CV). In particular, in the supervized learning setting Leave-One-Out (LOO) CV gives an almost unbiased estimate of the loss \cite{Sugiyama2007}. We develop Leave-One-Trajectory-Out (LOTO) CV, but unfortunately LOTO-CV is too computationally expensive for many practical applications.

 \noindent {\bf Efficient Cross-Validation:} We show how LOTO-CV can be efficiently implemented under the framework of Least Squares TD (LSTD($\lambda$) and Recursive LSTD($\lambda$)). Combining these ideas we propose \Algorithm\ (\Alg). While a na\"{i}ve implementation of LOTO-CV requires $O(kn)$ evaluations of LSTD, \Alg\ requires only $O(k)$ evaluations, where $n$ is the number of trajectories and $k = |\Lambda|$.

 Our experiments demonstrate that our proposed algorithm is effective at selecting $\lambda$ to minimize MSE. In addition, the experiments demonstrate that our proposed algorithm is significantly computationally faster than a na\"{i}ve implementation.

 \noindent {\bf Contributions:} The main contributions of this work are:
 \begin{enumerate}
   \item Formalize the $\lambda$ selection problem as finding the $\lambda$ value that leads to the smallest Mean Squared Value Error (MSVE),
   \item Develop LOTO-CV and propose using it to search the space of $\lambda$ values,
   \item Show how LOTO-CV can be implemented efficiently for LSTD,
   \item Introduce \Alg\ that is significantly computationally faster than the na\"{i}ve LOTO-CV implementation, and
   \item Prove that \Alg\ converges to the optimal hypothesis.
 \end{enumerate}

 \section{Background}

 Let $M = \langle \mathcal{S}, \mathcal{A}, P, r, \gamma \rangle$ be a Markov Decision Process (MDP) where $\mathcal{S}$ is a countable set of states, $\mathcal{A}$ is a finite set of actions, $P(s'|s, a)$ maps each state-action pair $(s, a) \in \mathcal{S} \times \mathcal{A}$ to the probability of transitioning to $s' \in \mathcal{S}$ in a single timestep, $r$ is an $|\mathcal{S}|$ dimensional vector mapping each state $s \in \mathcal{S}$ to a scalar reward, and $\gamma \in [0, 1]$ is the discount factor. We assume we are given a function $\phi : \mathcal{S} \rightarrow \mathbb{R}^d$ that maps each state to a $d$-dimensional vector, and we denote by $X = \phi(\mathcal{S})$ a $d \times |\mathcal{S}|$ dimensional matrix with one column for each state $s \in \mathcal{S}$.

 Let $\pi$ be a stochastic policy and denote by $\pi(a|s)$ the probability that the policy executes action $a \in \mathcal{A}$ from state $s \in \mathcal{S}$. Given a policy $\pi$, we can define the value function
 \begin{align}
   \nu^\pi &= \sum_{t=1}^{\infty} (\gamma P_\pi)^{t-1} r \enspace , \\
   &= r + \gamma P_\pi \nu^{\pi} \enspace ,
 \end{align}
 where $P_\pi(i,j) = \sum_{a \in \mathcal{A}} \pi(a|i) P(j|i,a)$. Note that $P_\pi$ is a $|\mathcal{S}| \times |\mathcal{S}|$ matrix where the $i^{\rm th}$ row is the probability distribution over next states, given that the agent is in state $i \in \mathcal{S}$.

 Given that $X^\top \theta \approx \nu^\pi$, we have that
 \begin{align}
   \nu^{\pi} &= r + \gamma P_\pi \nu^\pi \enspace , \nonumber \\
   \nu^{\pi} - \gamma P_\pi \nu^\pi &= r \enspace , \nonumber \\
   X^\top \theta - \gamma P_\pi X^\top \theta &\approx r \enspace , \nonumber && \textrm{Replace } \nu^\pi \textrm{ with } X^\top \theta . \\
   \underbrace{Z(X^\top - \gamma P_\pi X^\top)}_{\rm A} \theta &\approx \underbrace{Z r}_{\rm b} \enspace . && \times \textrm{ both sides by } Z .
 \end{align}
 where $Z$ is a $d \times |\mathcal{S}|$ matrix, which implies that $A$ is a $d \times d$ matrix and $b$ is a $d$ dimensional vector. Given $n \geq 1$ trajectories with length $H$\footnote{For clarity, we assume all trajectories have the same fixed length. The algorithms presented can easily be extended to handle variable length trajectories.}, this suggests the LSTD($\lambda$) algorithm \cite{Bradtke1996,Boyan2002}, which estimates
 \begin{align}
   A_\lambda &= \sum_{i=1}^n \sum_{t=1}^H z_{i,t}^{(\lambda)} w_{i,t}^\top \enspace , \label{eqn:lstd_amat} \\
   b_\lambda &= \sum_{i=1}^n \sum_{t=1}^H z_{i,t}^{(\lambda)} r_{i,t} \label{eqn:lstd_b} \enspace ,
 \end{align}
 where $z_{i,j}^{(\lambda)} = \sum_{t=1}^{j} (\lambda \gamma)^{j-t} x_{i,t}$, $w_{i,t} = \left( x_{i,t} - \gamma x_{i,t+1} \right)$, and $\lambda \in [0, 1]$. After estimating $A_\lambda$ and $b_\lambda$, LSTD solves for the parameters
 \begin{equation}
   \theta_\lambda = A_\lambda^{-1} b_\lambda \enspace . \label{eqn:lstd_solve}
 \end{equation}
 We will drop the subscript $\lambda$ when it is clear from context. The computational complexity of LSTD($\lambda$) is $O(d^3 + nHd^2)$, where the $O(d^3)$ term is due to solving for the inverse of $A_\lambda$ and the $O(nHd^2)$ term is the cost associated with building the $A$ matrix. We can further reduce the total computational complexity to $O(nHd^2)$ by using Recursive LSTD($\lambda$) \cite{Xu2002}, which we will refer to as RLSTD($\lambda$). Instead of computing $A_\lambda$ and solving for its inverse, RLSTD($\lambda$) recursively updates an estimate $\hat{A}_\lambda^{-1}$ of $A_\lambda^{-1}$ using the Sherman-Morrison formula \cite{Sherman1949}.
\begin{define}
  If $M$ is an invertable matrix and $u, v$ are column vectors such that $1 + v^\top M^{-1} u \neq 0$, then the {\bf Sherman-Morrison formula} is given by
  \begin{equation} \label{eq:sm_formula}
  \left( M + uv^\top \right)^{-1} = M^{-1} - \frac{M^{-1}uv^\top M^{-1}}{1 + v^\top M^{-1}u} \enspace .
\end{equation}
\end{define}
 RLSTD($\lambda$) updates $\hat{A}_\lambda^{-1}$ according to the following rule:
 \begin{equation}
   \hat{A}_\lambda^{-1} \leftarrow \left\{ \begin{array}{ll} \frac{1}{\rho} I_{d\times d} & \textrm{if } i = 0 \wedge t = 0 \\ \textrm{SM}(\hat{A}_\lambda^{-1}, z^{(\lambda)}_{i,t}, w_{i,t})  \end{array} \right. \enspace ,
 \end{equation}
 where $\rho > 0$, $I_{d \times d}$ is the $d \times d$ identity matrix, and SM is the Sherman-Morrison formula given by \eqref{eq:sm_formula} with $M = \hat{A}_\lambda^{-1}$, $u = z^{(\lambda)}_{i,t}$, and $v = w_{i, t}$.

 In the remainder of this paper, we focus on LSTD rather than RLSTD for (a) clarity, (b) because RLSTD has an additional initial variance parameter, and (c) because LSTD gives exact least squares solutions (while RLSTD's solution is approximate). Note, however, that similar approaches and analysis can be applied to RLSTD.

\section{Adapting the Timescale of LSTD}

  The parameter $\lambda$ effectively controls the timescale at which updates are performed. This induces a bias-variance trade-off, because longer timescales ($\lambda$ close to 1) tend to result in high variance, while shorter timescales ($\lambda$ close to 0) introduce bias. In this paper, the solution to this trade-off is the value of $\lambda \in \Lambda \subseteq [0, 1]$ that produces the parameters $\theta_\lambda$ that minimize Mean Squared Value Error (MSVE)
  \begin{equation}
    \| \nu^\pi - X^\top \theta_\lambda \|_\mu^2 = \sum_{s \in \mathcal{S}} \mu(s) \left( \nu^\pi(s) - \phi(s)^\top \theta_\lambda \right)^2 \enspace ,
  \end{equation}
  where $\mu(s)$ is a distribution over states.

  If $\Lambda$ is a finite set, a natural choice is to perform Leave-One-Out (LOO) Cross-Validation (CV) to select the $\lambda \in \Lambda$ that minimizes the MSE. Unlike the typical supervised learning setting, individual sampled transitions are correlated. So the LOO-CV errors are potentially biased. However, since trajectories are independent, we propose the use of Leave-One-Trajectory-Out (LOTO) CV.

Let $k = |\Lambda|$, then a na\"{i}ve implementation would perform LOTO-CV for each parameter in $\Lambda$. This would mean running LSTD $O(n)$ times for each parameter value in $\Lambda$. Thus the total time to run LOTO-CV for all $k$ parameter values is $O\left( kn \left[ d^3 + nHd^2 \right] \right)$. The na\"{i}ve implementation is slowed down significantly by the need to solve LSTD $O(n)$ times for each parameter value. We first decrease the computational cost associated with LOTO-CV for a single parameter value. Then we consider methods that reduce the cost associated with solving LSTD for $k$ different values of $\lambda$ rather than solving for each value separately.

 \subsection{Efficient Leave-One-Trajectory-Out CV}

 Fix a single value $\lambda \in [0, 1]$. We denote by
 \begin{align}
   C_{(i)} &= \sum_{j \neq i} \sum_{t=1}^H z_{j,t} \left( x_{j,t} - \gamma x_{j,t+1} \right)^\top \enspace , \\
   y_{(i)} &= \sum_{j \neq i} \sum_{t=1}^H z_{j,t} r_{j,t} \enspace , \textrm{ and} \\
   \theta_{(i)} &= C_{(i)}^{-1} y_{(i)} \enspace ,
 \end{align}
 where $\theta_{(i)}$ is the LSTD($\lambda$) solution computed without the $i^{\rm th}$ trajectory.

 The LOTO-CV error for the $i^{\rm th}$ trajectory is defined by
\begin{equation}
  [\ell]_i = \frac{1}{H} \sum\limits_{t=1}^H \left( x_{i,t}^\top \theta_{(i)} - \sum\limits_{j=t}^{H} \gamma^{j-t}r_{i,j} \right)^2 \enspace ,
\end{equation}
which is the Mean Squared Value Error (MSVE). Notice that the LOTO-CV error only depends on $\lambda$ through the computed parameters $\theta_{(i)}$. This is an important property because it allows us to compare this error for different choices of $\lambda$. Once the parameters $\theta_{(i)}$ are known, the LOTO-CV error for the $i^{\rm th}$ trajectory can be computed in $O(Hd)$ time.

 Since $\theta_{(i)} = C_{(i)}^{-1} y_{(i)}$, it is sufficient to derive $C_{(i)}^{-1}$ and $y_{(i)}$. Notice that $C_{(i)} = A - \sum\limits_{t=1}^{H} z_{i,t} \left( x_{i,t} - \gamma x_{i,t+1}  \right)^\top$ and $y_{(i)} = b - \sum\limits_{t=1}^H z_{i,t} r_{i, t}$. After deriving $A$ and $b$ via (\ref{eqn:lstd_amat}) and (\ref{eqn:lstd_b}), respectively, we can derive $y_{(i)}$ straightforwardly in $O(Hd)$ time. However, deriving $C_{(i)}^{-1}$ must be done more carefully. We first derive $A^{-1}$ and then update this matrix recursively using the Sherman-Morrison formula to remove each transition sample from the $i^{\rm th}$ trajectory.

\begin{algorithm}
  \caption{Recursive Sherman-Morrison (RSM)}
  \label{alg:rsm}
  \begin{algorithmic}[1]
    \REQUIRE $M$ an $d \times d$ matrix, $\mathcal{D} = \{(u_t, v_t)\}_{t=1}^{T}$ a collection of $2T$ $d$-dimensional column vectors.
    \STATE $\widetilde{M}_0 \leftarrow M$
    \FOR{$t = 1, 2, \dots, T$}
    \STATE $\widetilde{M}_t \leftarrow \widetilde{M}_{t-1} + \frac{\widetilde{M}_{t-1} u_tv_t^\top \widetilde{M}_{t-1}}{1 + v_t^\top \widetilde{M}_{t-1} u_t}$
    \ENDFOR
    \RETURN $\widetilde{C}_{T}$
  \end{algorithmic}
\end{algorithm}

We update $A^{-1}$ recursively for all $H$ transition samples from the $i^{\rm th}$ trajectory via
\begin{align*}
  C_{(i)}^{-1} &= \left( A - \sum\limits_{t=1}^{H} z_{i,t} (x_{i,t} - \gamma x_{i,t+1})^\top \right)^{-1} \enspace , \\
  &= \left( A + \sum_{t=1}^H z_{i,t} (\gamma x_{i,t+1} - x_{i,t})^\top \right)^{-1} \enspace , \\
  &= \left( A + \sum_{t=1}^H u_t v_t^\top \right)^{-1} \enspace ,
\end{align*}
where $u_t = z_{i,t}$ and $v_t = (\gamma x_{i,t+1} - x_{i,t})$. Now applying the Sherman-Morrison formula, we can obtain
\begin{equation}
  \widetilde{C}_t^{-1} = \left\{ \begin{array}{ll} A^{-1} & \textrm{if } t=0 \\ \widetilde{C}^{-1}_{t-1} - \frac{\widetilde{C}^{-1}_{t-1} u_t v_t^\top \widetilde{C}^{-1}_{t-1} }{ 1 + v_t^\top \widetilde{C}^{-1}_{t-1} u_t} & 1 \leq t \leq H \end{array} \right. \enspace ,
\end{equation}
which gives $\widetilde{C}_H^{-1} = C_{(i)}^{-1}$. Since the Sherman-Morison formula can be applied in $O(d^2)$ time, erasing the effect of all $H$ samples for the $i^{\rm th}$ trajectory can be done in $O(Hd^2)$ time.

Using this approach the cost of LSTD($\lambda$) + LOTO-CV is $O(d^3 + nHd^2)$, which is on the same order as running LSTD($\lambda$) alone. So computing the additional LOTO-CV errors is practically free.

\begin{algorithm}
  \caption{LSTD($\lambda$) + LOTO-CV}
  \label{alg:lstd_cv}
  \begin{algorithmic}[1]
    \REQUIRE $\mathcal{D} = \{ \tau_i = \langle x_{i,t}, r_{i,t}, x_{i,t+1} \rangle_{t=1}^H \}_{i=1}^n, \lambda \in [0, 1]$
    \STATE Compute $A$ (\ref{eqn:lstd_amat}) and $b$ (\ref{eqn:lstd_b}) from $\mathcal{D}$.
    \STATE $\theta \leftarrow A^{-1} b$ \COMMENT{Compute $A^{-1}$ and solve for $\theta$.}
    \STATE $\ell \leftarrow {\bf 0}$ \COMMENT{An $n$-dimensional column vector.}
    \FOR[Leave-One-Trajectory-Out]{$i = 1, 2, \dots, n$}
    \STATE $C_{(i)}^{-1} \leftarrow {\rm RSM}(A^{-1}, \{ z_{i,t}, (\gamma x_{i,t+1} - x_{i,t}) \}_{t=1}^H)$
    \STATE $\theta_{(i)} \leftarrow C_{(i)}^{-1} y$
    \STATE $[\ell ]_i \leftarrow \frac{1}{H} \sum\limits_{t=1}^H \left( x_{i,t}^\top \theta_{(i)} - \sum\limits_{j=t}^{H} \gamma^{j-t}r_{i,j}  \right)^2$
    \ENDFOR
    \RETURN $\theta$ \COMMENT{LSTD($\lambda$) solution.}, $\ell$ \COMMENT{LOTO errors.}
 \end{algorithmic}
\end{algorithm}

\subsection{Solving LSTD for $k = |\Lambda|$ Timescale Parameter Values}
Let $\widehat{X}_{(i)}$ be a $d \times H$ matrix where the columns are the state observation vectors at timesteps $t = 1, 2, \dots, H$ during the $i^{\rm th}$ trajectory (with the last state removed) and $\widehat{W}_{(i)}$ be a $d \times H$ matrix where the columns are the next state observation vectors at timesteps $t = 2, 3, \dots, H+1$ during the $i^{\rm th}$ trajectory (with the first state observation removed). We define $\widehat{X} = \langle \widehat{X}_{(1)}, \widehat{X}_{(2)}, \dots, \widehat{X}_{(n)} \rangle$ and $\widehat{W} = \langle \widehat{W}_{(1)}, \widehat{W}_{(2)}, \dots, \widehat{W}_{(n)} \rangle$, which are both $d \times nH$ matricies.

\begin{align}
  A_\lambda &= \widehat{Z} ( \widehat{X} - \gamma \widehat{W})^\top \enspace , \nonumber \\
  &= (\widehat{Z} - \widehat{X} + \widehat{X}) (\widehat{X} - \gamma \widehat{W})^\top \enspace , \nonumber \\
  &= (\widehat{Z} - \widehat{X})(\widehat{X} - \gamma \widehat{W})^\top + A_0 \enspace , \nonumber \\
  &= \sum\limits_{i=1}^n \sum\limits_{t=1}^H u_{i,t} v_{i,t}^\top + A_0 \enspace , \label{eqn:amat_lambda_to_zero}
\end{align}
where $u_{i,t} = (z^{(\lambda)}_{i,t} - x_{i,t})$ and $v_{i,t} = (x_{i,t} - \gamma x_{i,t+1})$.

By applying the Sherman-Morrison formula recursively with $u_{i,t} = (z_{i,t}-x_{i,t})$ and $v_{i,t}=(x_{i,t}-\gamma x_{i,t+1})$, we can obtain $A_\lambda^{-1}$ in $O(nHd^2)$ time and then obtain $\theta_\lambda$ in $O(d^2)$ time. Thus, the total running time for LSTD with $k$ timescale parameter values is $O(d^3 + knHd^2)$.

\subsection{Proposed Algorithm: \Alg}

\begin{algorithm}
  \caption{\Alg}
  \label{alg:main}
  \begin{algorithmic}[1]
    \REQUIRE $\mathcal{D} = \{ \tau_i = \langle x_{i,t}, r_{i,t}, x_{i,t+1} \rangle_{t=1}^H \}_{i=1}^n, \Lambda \subset [0, 1]$
    \STATE Compute $A_0$ (\ref{eqn:lstd_amat}) and $b_0$ (\ref{eqn:lstd_b}) from $\mathcal{D}$.
    \STATE Compute $A_0^{-1}$
    \STATE $\ell \leftarrow {\bf 0}$ \COMMENT{A $|\Lambda|$ vector.}
    \FOR{$\lambda \in \Lambda$}
    \STATE $A_\lambda^{-1} \leftarrow {\rm RSM}(A_0, \{ (u_j, v_j) \}_{j=1}^{nH})$ where $u_j = (z_{i,t}^{(\lambda)} - x_{i,t})$ and $v_j = (x_{i,t} - \gamma x_{i,t+1})$.
    \FOR{$i = 1, 2, \dots, n$}
    \STATE $C^{-1}_{(i)} \leftarrow RSM(A_\lambda^{-1}, \{ z_{i,t}^{(\lambda)}, (\gamma x_{i,t+1} - x_{i,t}) \}_{t=1}^H)$
    \STATE $\theta_{(i)} \leftarrow C_{(i)}^{-1}y_{(i)}$
    \STATE $[\ell]_\lambda \leftarrow [\ell]_\lambda + \frac{1}{H} \sum\limits_{t=1}^H \left( x_{i,t}^\top \theta_{(i)} - \sum\limits_{j=t}^H \gamma^{j-t}r_{i,j} \right)^2$
    \ENDFOR
    \ENDFOR
    \STATE $\lambda^* \leftarrow \arg \min_{\lambda \in \Lambda} [\ell]_\lambda$
    \RETURN $\theta_{\lambda^*}$
  \end{algorithmic}
\end{algorithm}

We combine the approaches from the previous two subsections to define \Algorithm\ (\Alg). The pseudo-code is given in Algorithm \ref{alg:main}. \Alg\ takes as input a set of $n\geq 2$ trajectories and a finite set of values $\Lambda$ in $[0, 1]$. $\Lambda$ is the set of candidate $\lambda$ values.

\begin{theorem} {\bf (Agnostic Consistency)} \label{thm:consistency}
Let $\Lambda \subseteq [0, 1]$, $\mathcal{D}_n$ be a dataset of $n \geq 2$ trajectories generated by following the policy $\pi$ in an MDP $M$ with initial state distribution $\mu_0$. If $1 \in \Lambda$, then as $n \rightarrow \infty$,
\begin{equation}
\lim_{n \rightarrow \infty} \| \nu^\pi - \mathcal{A}(\mathcal{D}_n, \Lambda)^\top \phi(\mathcal{S})\|_\mu - \min_{\theta \in \mathbb{R}^d}\| \nu^\pi - \theta^\top \phi(\mathcal{S}) \|_\mu = 0 \enspace ,
\end{equation}
where $\mathcal{A}$ is the proposed algorithm \Alg\ which maps from a dataset and $\Lambda$ to a vector in $\mathbb{R}^d$ and $\mu(s) = \frac{1}{H+1} \mu_0(s) + \frac{1}{H+1}\sum_{t=1}^H \sum_{s' \in S} \left( P^\pi \right)^t (s|s')\mu_{0}(s')$.
\end{theorem}
Theorem \ref{thm:consistency} says that in the limit \Alg\ converges to the best hypothesis. The proof of Theorem \ref{thm:consistency} is given in the appendix.

\section{Experiments}

We compared the MSVE and computation time of \Alg\ against a na\"{i}ve implementation of LOTO-CV (which we refer to as Na\"{i}veCV+LSTD) in three domains: random walk, 2048, and Mountain Car. As a baseline, we compared these algorithms to LSTD and RLSTD with the best and worst fixed choices of $\lambda$ in hindsight, which we denote LSTD(best), LSTD(worst), RLSTD(best), and RLSTD(worst). In all of our experiments, we generated 80 independent trials. In the random walk and 2048 domains we set the discount factor to $\gamma = 0.95$. For the mountain car domain we set the discount factor to $\gamma = 1$.

\begin{figure*}
  \centering
  \begin{tabular}{ccc}
    \includegraphics[width=0.32\textwidth]{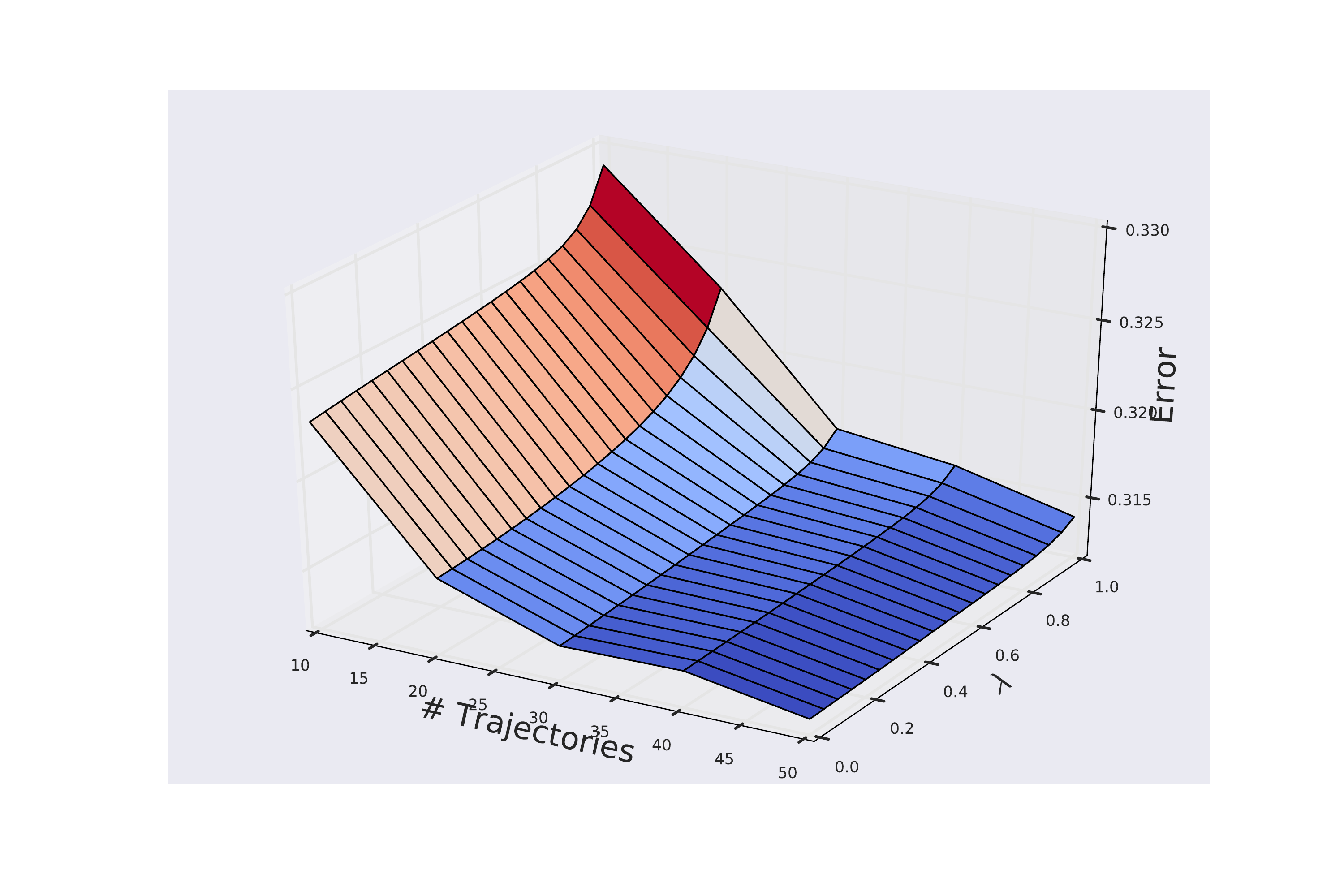} &
    \includegraphics[width=0.32\textwidth]{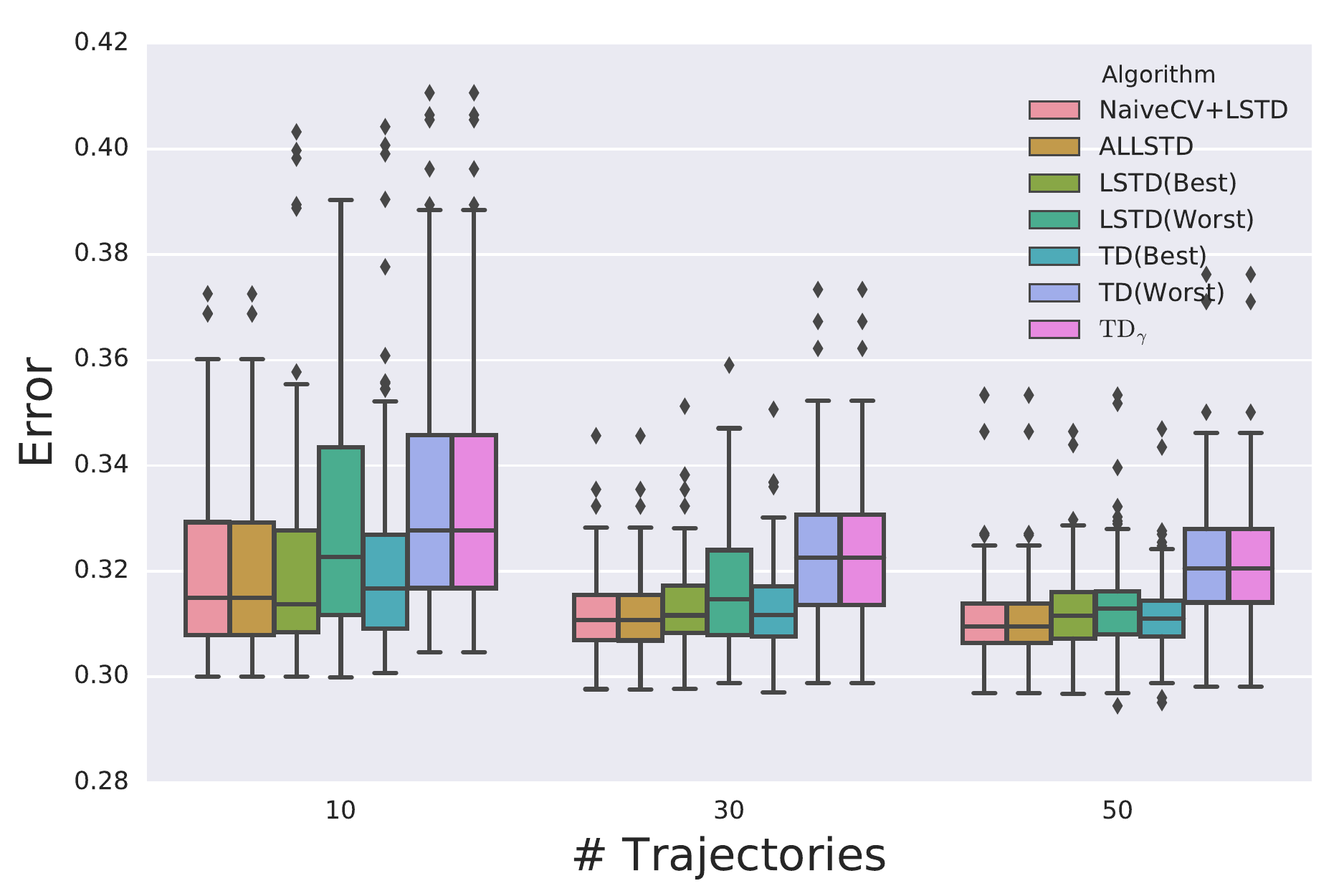} &
    \includegraphics[width=0.32\textwidth]{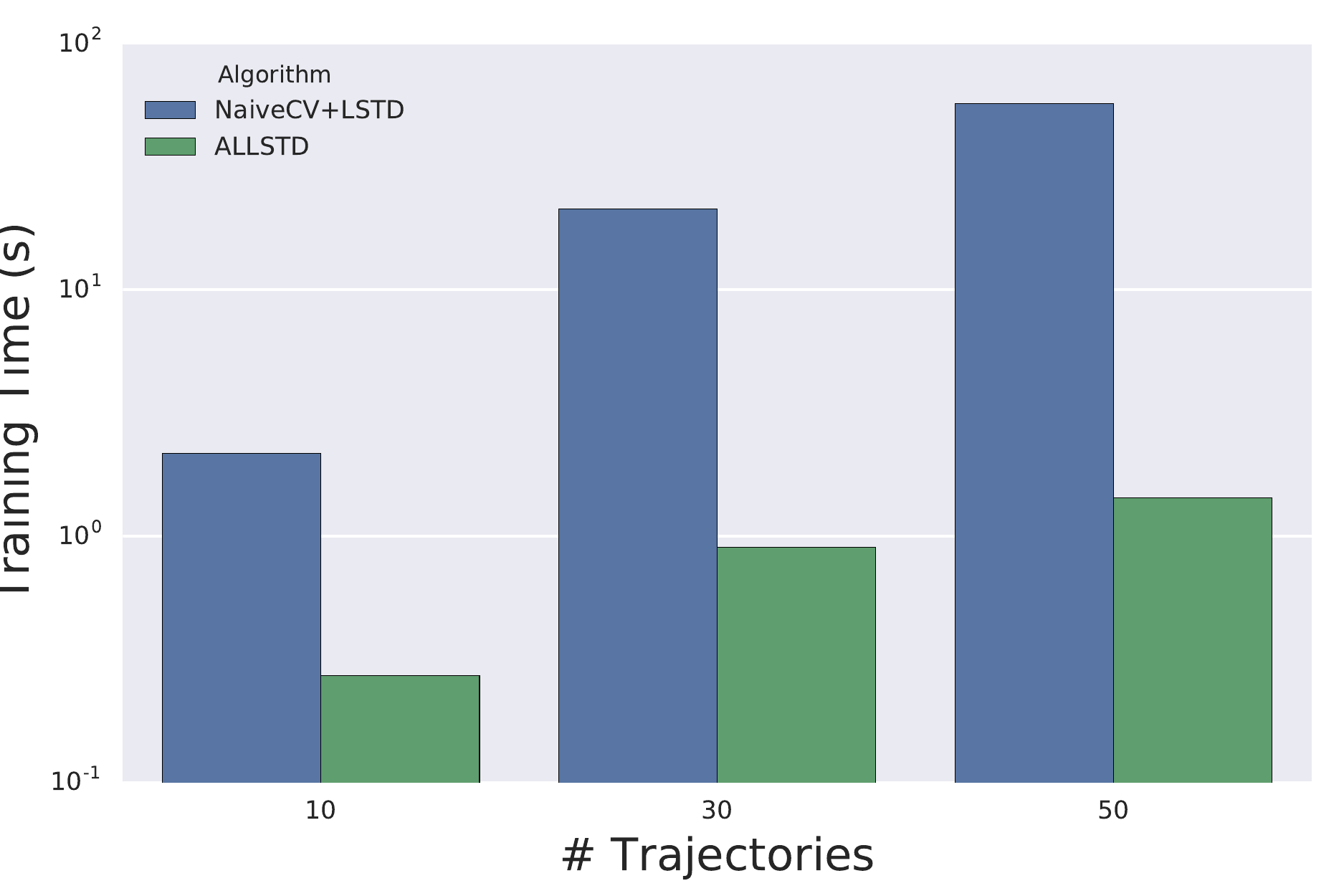} \\
    ($a$) & ($b$) & ($c$)
  \end{tabular}
	\caption{Random Walk domain: ($a$) The relationship between $\lambda$ and amount of training data (i.e., \# trajectories). ($b$) root MSVE as the \# trajectories is varied. \Alg\ achieves the same performance as Na\"{i}veCV+LSTD. ($c$) Training time in seconds as the \# trajectories is varied. \Alg\ is approximately and order of magnitude faster than Na\"{i}veCV+LSTD.}
  \label{fig:random_walk}
\end{figure*}

\begin{figure*}
  \centering
  \begin{tabular}{ccc}
    \includegraphics[width=0.32\textwidth]{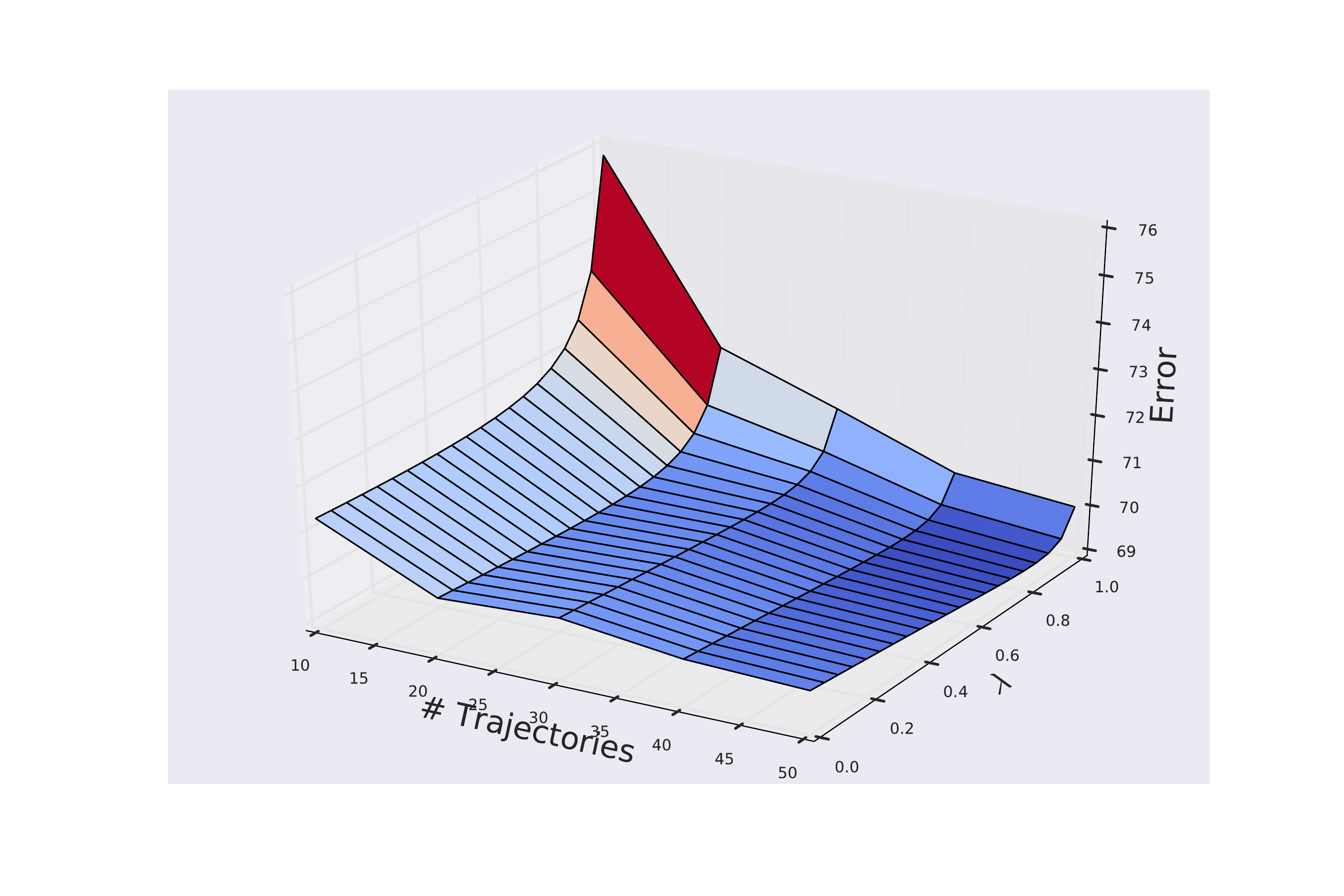} &
    \includegraphics[width=0.32\textwidth]{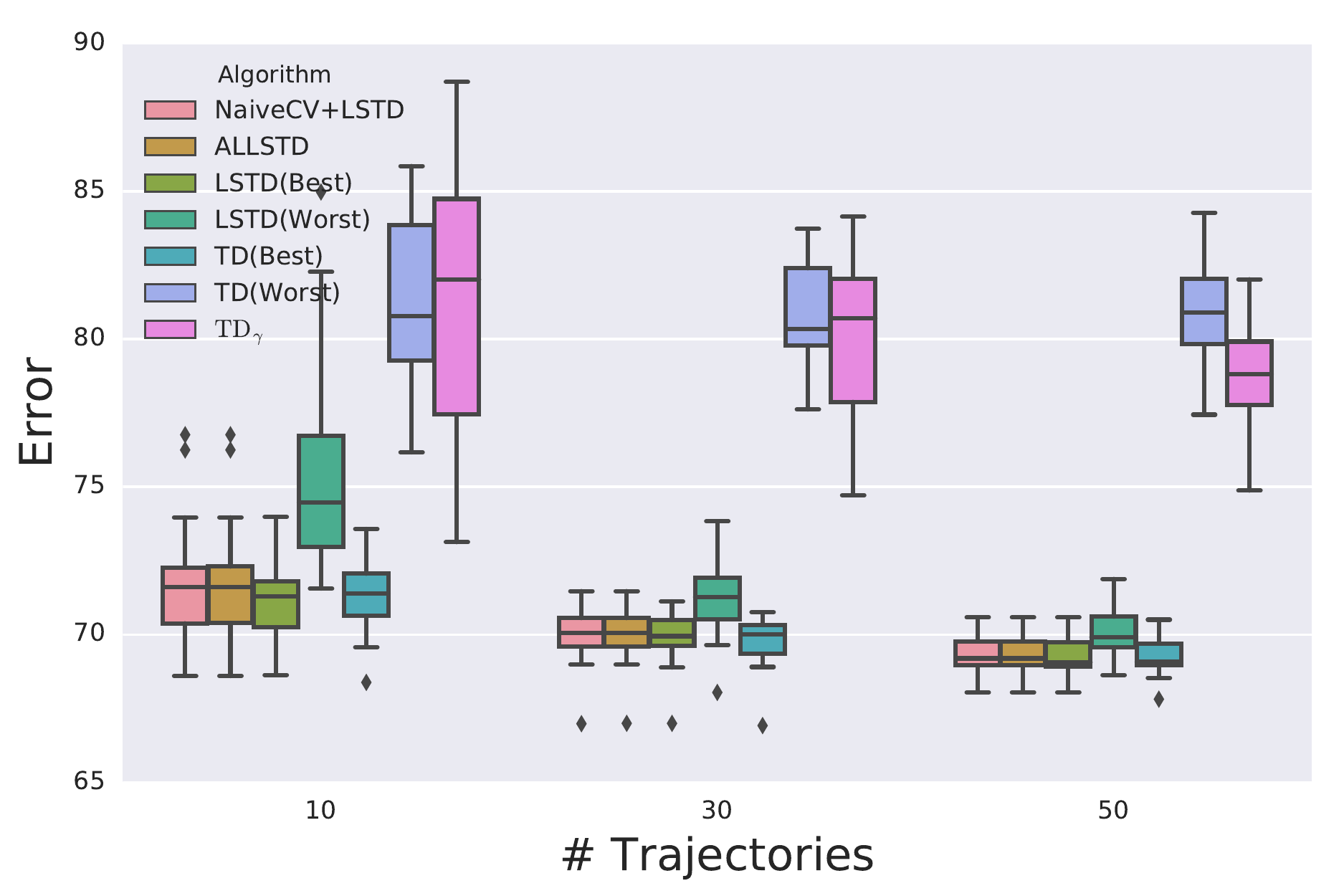} &
    \includegraphics[width=0.32\textwidth]{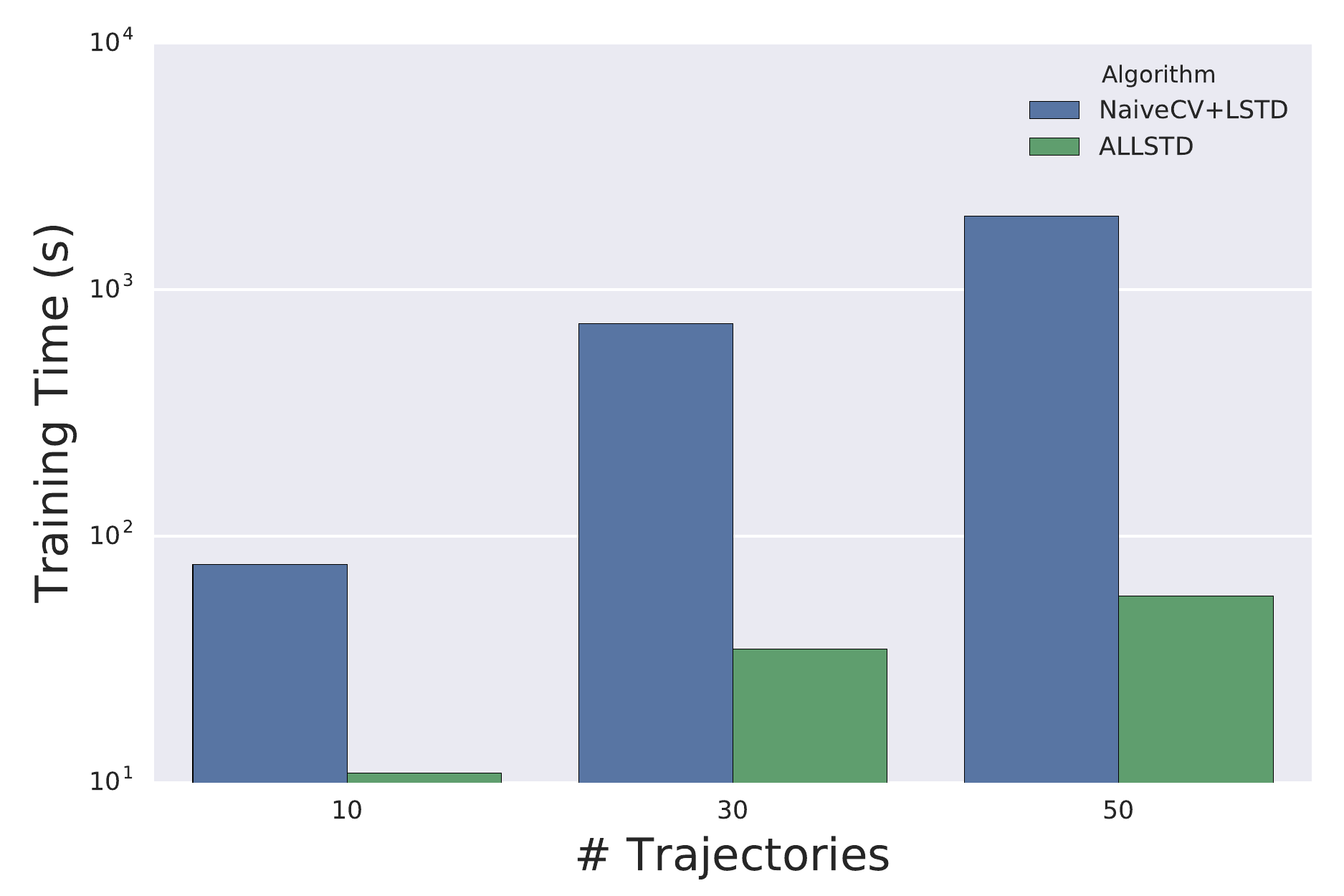} \\
    ($a$) & ($b$) & ($c$)
  \end{tabular}
  \caption{2048: ($a$) The relationship between $\lambda$ and amount of training data (i.e., \# trajectories). ($b$) root MSVE as the \# trajectories is varied. \Alg\ achieves the same performance as Na\"{i}veCV+LSTD. ($c$) Training time in seconds as the \# trajectories is varied. \Alg\ is approximately and order of magnitude faster than Na\"{i}veCV+LSTD.}
  \label{fig:2048}
\end{figure*}

\begin{figure*}
  \centering
  \begin{tabular}{ccc}
    \includegraphics[width=0.32\textwidth]{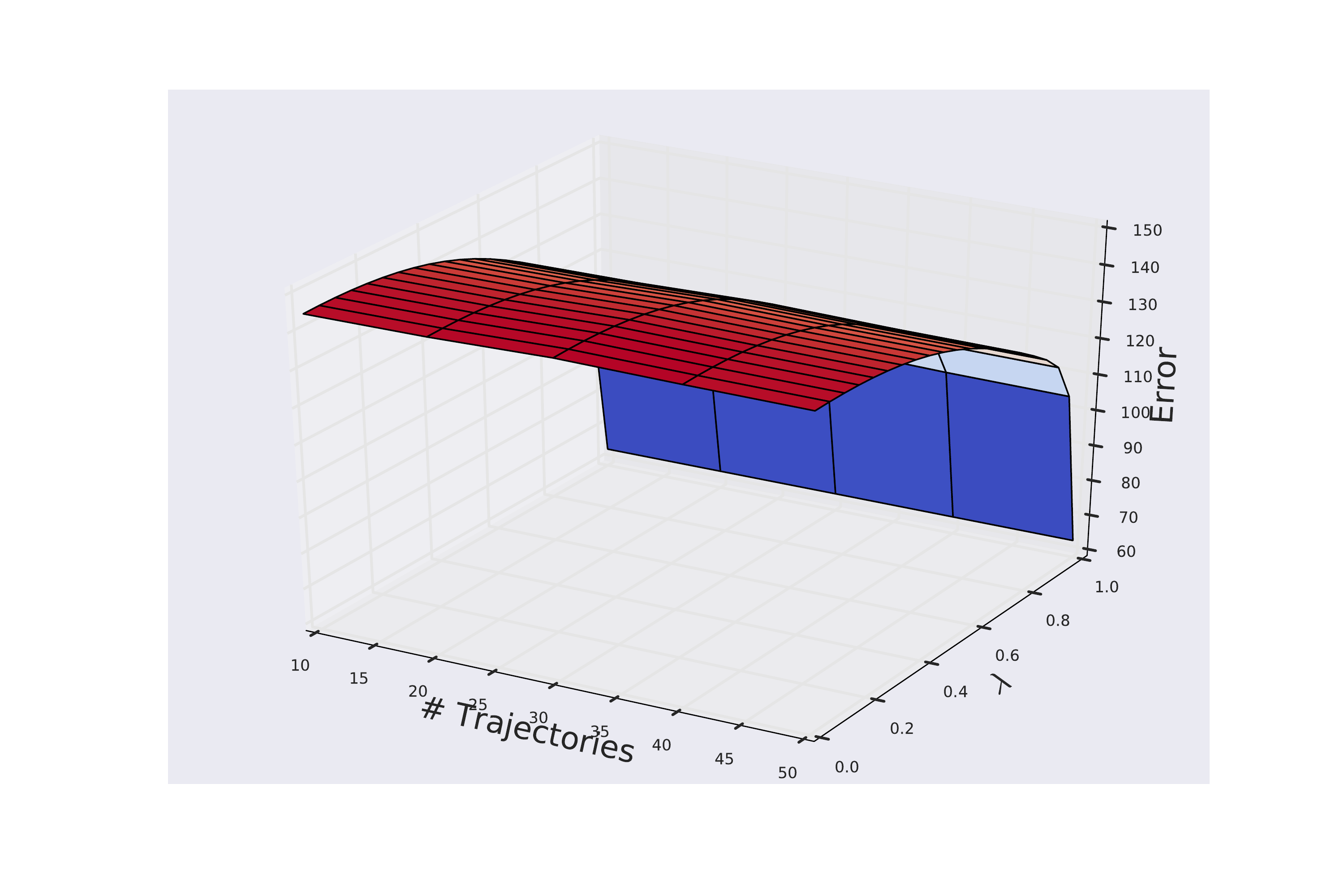} &
    \includegraphics[width=0.32\textwidth]{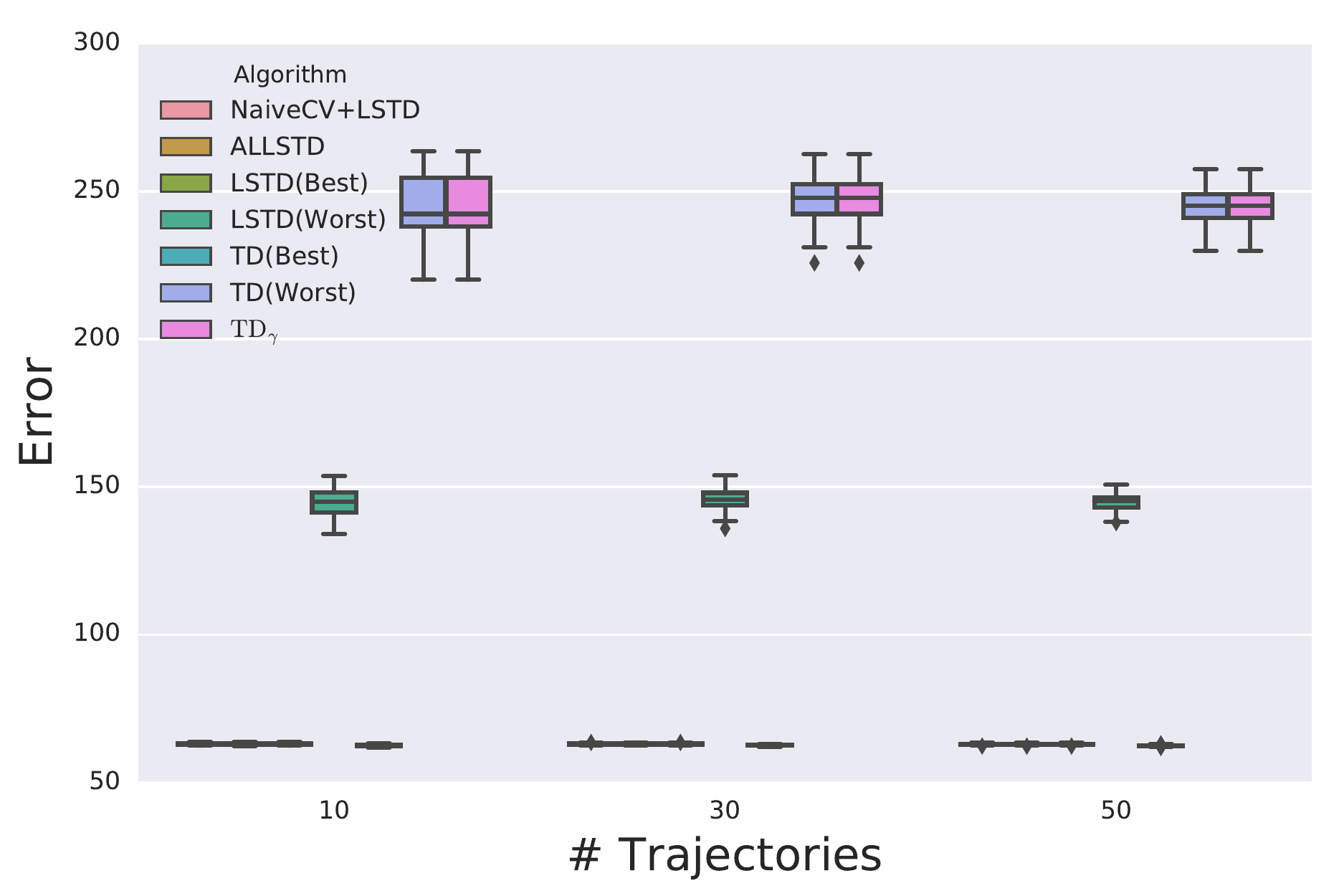} &
    \includegraphics[width=0.32\textwidth]{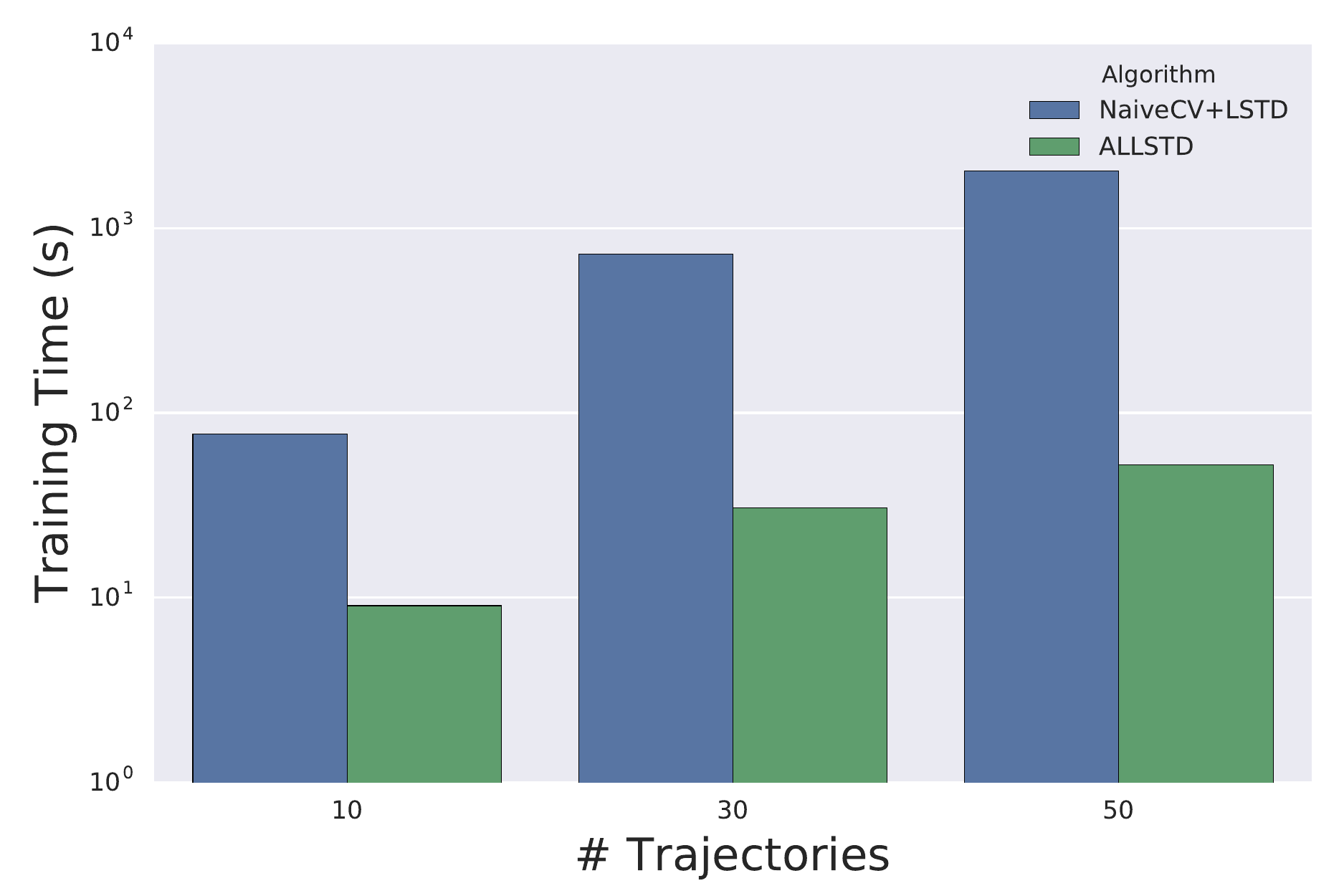} \\
    ($a$) & ($b$) & ($c$)
  \end{tabular}
  \caption{Mountain Car domain: ($a$) The relationship between $\lambda$ and amount of training data (i.e., \# trajectories). ($b$) root MSVE as the \# trajectories is varied. \Alg\ achieves the same performance as Na\"{i}veCV+LSTD. ($c$) Training time in seconds as the \# trajectories is varied. \Alg\ is approximately and order of magnitude faster than Na\"{i}veCV+LSTD.}
  \label{fig:mountain_car}
\end{figure*}

\subsection{Domain: Random Walk}
The random walk domain \cite{Sutton1998} is a chain of five states organized from left to right. The agent always starts in the middle state and can move left or right at each state except for the leftmost and rightmost states, which are absorbing. The agent receives a reward of 0 unless it enters the rightmost state where it receives a reward of $+1$ and the episode terminates.

{\bf Policy:} The policy used to generate the trajectories was the uniform random policy over two actions: left and right.
{\bf Features:} Each state was encoded using a 1-hot representation in a 5-dimensional feature vector. Thus, the value function was exactly representable.

Figure \ref{fig:random_walk}$a$ shows the root MSVE as a function of $\lambda$ and \# trajectories. Notice that $\lambda < 1$ results in lower error, but the difference between $\lambda < 1$ and $\lambda = 1$ decreases as the \# trajectories grows.

Figure \ref{fig:random_walk}$b$ compares root MSVE in the random walk domain. Notice that \Alg\ and Na\"{i}ve\Alg\ achieve roughly the same error level as LSTD(best) and RLSTD(best). While this domain has a narrow gap between the performance of LSTD(best) and LSTD(worst), \Alg\ and Na\"{i}ve\Alg\ achieve performance comparable to LSTD(best).

Figure \ref{fig:random_walk}$c$ compares the average execution time of each algorithm in seconds (on a log scale). $k \times$ LSTD and $k \times$ RLSTD are simply the time required to compute LSTD and RLSTD for $k$ different $\lambda$ values, respectively. They are shown for reference and do not actually make a decision about which $\lambda$ value to use. \Alg\ is significantly faster than Na\"{i}ve\Alg\ and takes roughly the same computational time as solving LSTD for $k$ different $\lambda$ values.

\subsection{Domain: 2048}
2048 is a game played on a $4 \times 4$ board of tiles. Tiles are either empty or assigned a positive number. Tiles with larger numbers can be acquired by merging tiles with the same number. The immediate reward is the sum of merged tile numbers.

{\bf Policy:} The policy used to generate trajectories was a uniform random policy over the four actions: up, down, left, and right.
{\bf Features:} Each state was represented by a 16-dimensional vector where the value was taken as the value of the corresponding tile and 0 was used as the value for empty tiles. This linear space was not rich enough to capture the true value function.

Figure \ref{fig:2048}$a$ shows the root MSVE as a function of $\lambda$ and \# trajectories. Similar to the random walk domain, $\lambda < 1$ results in lower error.

Figure \ref{fig:2048}$b$ compares the root MSVE in 2048. Again \Alg\ and Na\"{i}ve\Alg\ achieve roughly the same error level as LSTD(best) and RLSTD(best) and perform significantly better than LSTD(worst) and RLSTD(worst) for a small number of trajectories.

Figure \ref{fig:2048}$c$ compares the average execution time of each algorithm in seconds (on a log scale). Again, \Alg\ is significantly faster than Na\"{i}ve\Alg.

\subsection{Domain: Mountain Car}
The mountain car domain \cite{Sutton1998} requires moving a car back and forth to build up enough speed to drive to the top of a hill. There are three actions: forward, neutral, and reverse.

{\bf Policy:} The policy generate the data sampled one of the three actions with uniform probability 25\% of the time. On the remaining 75\% of the time the forward action was selected if
\begin{equation}
  \dot{x} > 0.025 x + 0.01 \enspace ,
\end{equation}
and the reverse action was selected otherwise, where $x$ represents the location of the car and $\dot{x}$ represents the velocity of the car.
{\bf Features:} The feature space was a 2-dimensional vector where the first element was the location of the car and the second element was the velocity of the car. Thus, the linear space was not rich enough to capture the true value function.

Figure \ref{fig:mountain_car}$a$ shows the root MSVE as a function of $\lambda$ and \# trajectories. Unlike the previous two domains, $\lambda = 1$ achieves the smallest error even with a small \# trajectories. This difference is likely because of the poor feature representation used, which favors the Monte-Carlo return \cite{Bertsekas1996,Tagorti2015}.

Figure \ref{fig:mountain_car}$b$ compares the root MSVE in the mountain car domain. Because of the poor feature representation, the difference between LSTD(best) and LSTD(worst) is large. \Alg\ and Na\"{i}ve\Alg\ again achieve roughly the same performance as LSTD(best) and RLSTD(best).

Figure \ref{fig:mountain_car}$c$ shows that average execution time for \Alg\ is significantly shorter than for Na\"{i}ve\Alg.

\section{Related Work}

We have introduced \Alg\ to efficiently select $\lambda$ to minimize root MSVE in a data-driven way. The most similar approach to \Alg\ is found in the work of \citeauthor{Downey2010}, which introduces a Bayesian model averaging approach to update $\lambda$. However, this approach is not comparable to \Alg\ because it is not clear how it can be extended to domains where function approximation is required to estimate the value.

\citeauthor{Konidaris2011} and \citeauthor{Thomas2015} introduce $\gamma$-returns and $\Omega$-returns, respectively, which offer alternative weightings of the $t$-step returns. However, these approaches were designed to estimate the value of a single point rather than a value function. Furthermore, they assume that the bias introduced by $t$-step returns is $0$. \citeauthor{Thomas2016} introduce the MAGIC algorithm that attempts to account for the bias of the $t$-step returns, but this algorithm is still only designed to estimate the value of a point. \Alg\ is designed to estimate a value function in a data-driven way to minimize root MSVE.

\citeauthor{White2016} introduce the $\lambda$-greedy algorithm for adapting $\lambda$ per-state based on estimating the bias and variance. However, an approximation of the bias and variance is needed for each state to apply $\lambda$-greedy. Approximating these values accurately is equivalent to solving our original policy evaluation problem, and the approach suggested in the work of \citeauthor{White2016} introduces several additional parameters. \Alg, on the other hand, is a parameter free algorithm. Furthermore, none of these previous approaches suggest using LOTO-CV to tune $\lambda$ or show how LOTO-CV can be efficiently implemented under the LSTD family of algorithms.

\section{Discussion}

While we have focused on on-policy evaluation, the bias-variance trade-off controlled by $\lambda$ is even more extreme in off-policy evaluation problems. Thus, an interesting area of future work would be applying \Alg\ to off-policy evaluation \cite{White2016,Thomas2016}. It may be possible to identify good values of $\lambda$ without evaluating all trajectories. A bandit-like algorithm could be applied to determine how many trajectories to use to evaluate different values of $\lambda$. It is also interesting to note that our efficient cross-validation trick could be used to tune other parameters, such as a parameter controlling $L_2$ regularization.

In this paper, we have focused on selecting a single global $\lambda$ value, however, it may be possible to further reduce estimation error by learning $\lambda$ values that are specialized to different regions of the state space \cite{Downey2010,White2016}. Adapting $\lambda$ to different regions of the state-space is challenging because increases the search space, but it identifying good values of $\lambda$ could improve prediction accuracy in regions of the state space with high variance or little data.

\bibliographystyle{apalike}
\bibliography{attd}

\appendix
\appendixpage

\section{Agnostic Consistency of \Alg}

\begin{theorem} {\bf (Agnostic Consistency)} \label{thm:consistency}
Let $\Lambda \subseteq [0, 1]$, $\mathcal{D}_n$ be a dataset of $n \geq 2$ trajectories generated by following the policy $\pi$ in an MDP $M$ with initial state distribution $\mu_0$. If $1 \in \Lambda$, then as $n \rightarrow \infty$,
\begin{equation}
\lim_{n \rightarrow \infty} \| \nu^\pi - \mathcal{A}(\mathcal{D}_n, \Lambda)^\top \phi(\mathcal{S})\|_\mu - \min_{\theta \in \mathbb{R}^d}\| \nu^\pi - \theta^\top \phi(\mathcal{S}) \|_\mu = 0 \enspace ,
\end{equation}
where $\mathcal{A}$ is the proposed algorithm \Alg\ which maps from a dataset and $\Lambda$ to a vector in $\mathbb{R}^d$ and $\mu(s) = \frac{1}{H+1} \mu_0(s) + \frac{1}{H+1}\sum_{t=1}^H \sum_{s' \in S} \left( P^\pi \right)^t (s|s')\mu_{0}(s')$.
\end{theorem}
Theorem \ref{thm:consistency} says that in the limit \Alg\ converges to the best hypothesis.
\begin{proof}
  \Alg\ essentially executes LOTO-CV+LSTD($\lambda$) for $|\Lambda|$ parameter values and compares the scores returned. Then it returns the LSTD($\lambda$) solution for the $\lambda$ value with the lowest score. So it is sufficient to show that
  \begin{enumerate}
    \item the scores converge to the expected MSVE for each $\lambda \in \Lambda$, and
    \item that LSTD($\lambda=1$) converges to $\theta^* \in \arg \min_{\theta \in \mathbb{R}^d} \| \nu^\pi - \theta^\top \phi(\mathcal{S}) \|_\mu$.
  \end{enumerate}
  % The LOTO-CV scores converge
  The first follows by an application of the law of large numbers. Since each trajectory is independent, the MSVE for each $\lambda$ will converge almost surelyto the expected MSVE.
  % LSTD(1) converges
  The second follows from the fact that LSTD($\lambda=1$) is equivalent to linear regression against the Monte-Carlo returns \cite{Boyan2002}. Notice that the distribution $\mu$ is simply an average over the distributions encountered at each timestep.

\end{proof}

\end{document}